\documentclass[conference]{IEEEtran}
\IEEEoverridecommandlockouts
\usepackage{cite}
\usepackage{amsmath,amssymb,amsfonts}
\usepackage{algorithmic}
\usepackage{graphicx}
\usepackage{textcomp}
\usepackage{xcolor}
\usepackage{amsmath}
\usepackage{comment}
\usepackage[linesnumbered,ruled,vlined]{algorithm2e}
\usepackage{subcaption}
\usepackage{enumitem}
\usepackage{hyperref}
\usepackage{graphicx}
\usepackage{tabularx}
\usepackage{array}
\usepackage{multirow}
\usepackage{tabularx}
\usepackage{enumitem}  

\newcolumntype{C}[1]{>{\centering\arraybackslash}p{#1}}
\newcolumntype{M}[1]{>{\centering\arraybackslash}m{#1}}

\def\BibTeX{{\rm B\kern-.05em{\sc i\kern-.025em b}\kern-.08em
    T\kern-.1667em\lower.7ex\hbox{E}\kern-.125emX}}
\begin{document}

\title{Building Multi-Agent Copilot towards Autonomous Agricultural Data Management and Analysis
\thanks{}
}

\author{

\IEEEauthorblockN{Yu Pan$^1$, Jianxin Sun$^{2,4}$, Hongfeng Yu$^{2,4}$, Joe Luck$^1$, Geng Bai$^5$, Nipuna Chamara$^1$, Yufeng Ge$^1$,  Tala Awada$^3$}
\IEEEauthorblockA{$^1$\textit{Department of Biological System Engineering, University of Nebraska-Lincoln, Lincoln, NE, USA} \\
$^2$\textit{School of Computing, University of Nebraska-Lincoln, Lincoln, NE, USA}\\
$^3$\textit{School of Natural Resources, University of Nebraska-Lincoln, Lincoln, NE, USA}\\
$^4$\textit{Holland Computing Center, University of Nebraska-Lincoln, Lincoln, NE, USA}\\
$^5$\textit{Department of Biological and Agricultural Engineering, North Carolina State University, Raleigh, NC,USA}
}
}

\maketitle

\begin{abstract}
The ubiquity of sensors and IoT devices has led to an explosion in data availability in modern agriculture. The large volume and heterogeneity of the data, together with the complexity of data processing requirements, pose huge obstacles for achieving the principles of Findable, Accessible, Interoperable, and Reusable (FAIR). Current data management and analysis paradigms are to large extent traditional, in which data collecting, curating, integration, loading, storing, sharing and analyzing still involve too much human effort and know-how. The experts, researchers and the farm operators need to understand the data and the whole process of data management pipeline to make fully use of the data. The essential problem of the traditional paradigm is the lack of a layer of orchestrational intelligence which can understand, organize and coordinate the data processing utilities to maximize data management and analysis outcome. The emerging reasoning and tool mastering abilities of large language models (LLM) make it a potentially good fit to this position, which helps a shift from the traditional user-driven paradigm to AI-driven paradigm. In this paper, we propose and explore the idea of a LLM based copilot for autonomous agricultural data management and analysis. Based on our previously developed platform of Agricultural Data Management and Analytics (ADMA), we build a proof-of-concept multi-agent system called ADMA Copilot, which can understand user's intent, makes plans for data processing pipeline and accomplishes tasks automatically, in which three agents: a LLM based controller, an input formatter and an output formatter collaborate together. Different from existing LLM based solutions, by defining a meta-program graph, our work decouples control flow and data flow to enhance the predictability of the behaviour of the agents. Experiments demonstrates the intelligence, autonomy, efficacy, efficiency, extensibility, flexibility and privacy of our system. Comparison is also made between ours and existing systems to show the superiority and potential of our system. 
\end{abstract}

\begin{IEEEkeywords}
Agricultural Data Management, Multi-Agent, Copilot, FAIR principles, Large Language Model, Autonomous, Paradigm shift
\end{IEEEkeywords}

\section{Introduction}
The core topics of modern agriculture is to meet the escalating needs for food, fuel, feed, and fiber in the era of climate change, limited natural resources and intricate dynamics of global ecosystems. To solve this challenge, data-informed precision agriculture is not only required but also necessary, in which the deployment of sensors, Internet of Things (IoT) devices \cite{chamara2022ag}, and advanced instrumentation \cite{bai2019nu} has enabled data collection in various modal, dimension, format, and resolution. While agricultural data has the potential to transform agricultural practice by providing valuable insights into crop productivity, soil quality, and sustainable farming techniques, in the meanwhile, the large volume and heterogeneity, together with the complexity of data processing requirements, pose huge obstacles for for achieving the principles of Findable, Accessible, Interoperable, and Reusable (FAIR), which in turn result in in-efficient utilization and exploration of the collected data.

There are existing efforts trying to revolutionize current practices of agriculture data management  \cite{swetnam2023cyverse,senay2022big,jouini2021evaluation,lebauer2020terra,pan2023transforming}, by implementing different data management platforms and technologies. Nevertheless, current practice of agricultural data management and analysis is to large extent traditional in the sense that the whole data management pipeline, including data collecting, curating, integration, loading, storing, sharing and analyzing still involve too much human effort and know-how. The experts, researchers or the farm operators make every tiny decisions from where to put a file, to which analysis tool to use. In this process, they need to understand the details of the data and tools at their hands, including location, semantics and format of data, and the location, semantics and the interface of the relevant data processing tools to make fully use of the data. This human-driven paradigm incurs high cost of training and makes reuse of the data and duplication of the data processing pipeline difficult. The essential reason of the challenges faced by the traditional practice of agricultural data management lies in the fact that there's a lack of a layer of orchestrational intelligence which can understand, organize and coordinate each phase of the data processing pipeline to maximize the outcome. 

\begin{figure}[t]
    \centering
    \begin{subfigure}[b]{0.9\linewidth}
        \centering
        \includegraphics[width=\linewidth]{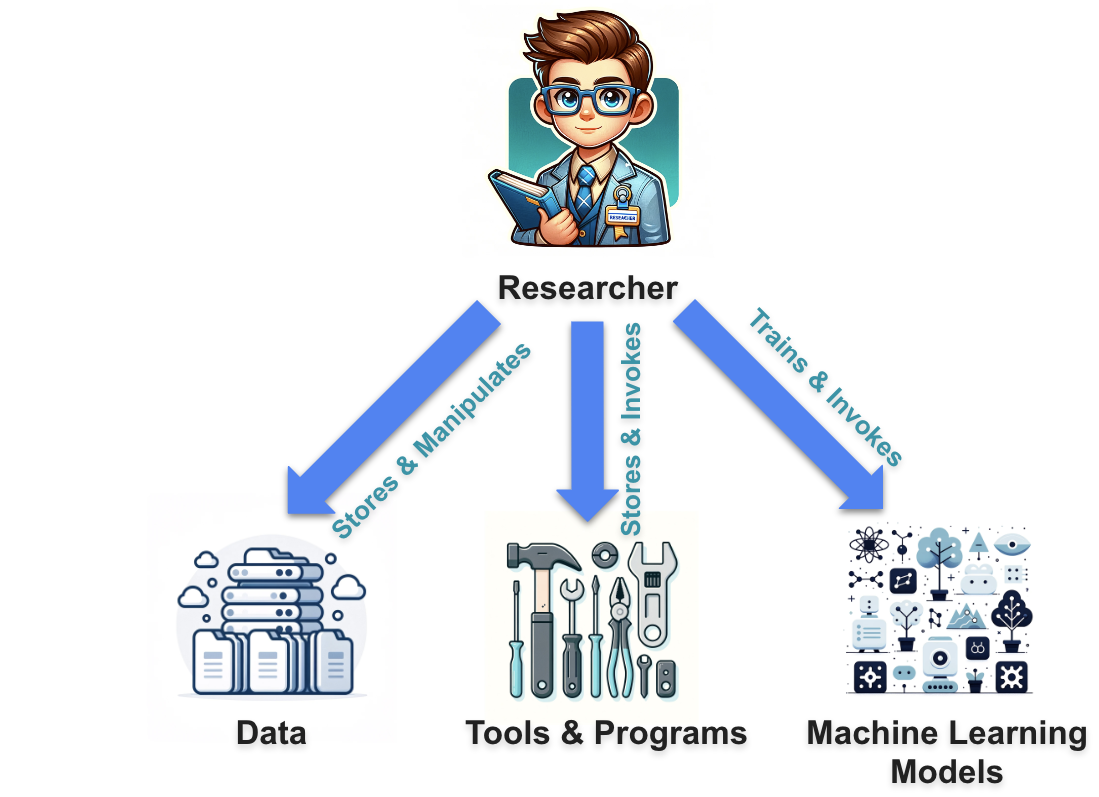}
        \caption{User-Driven Paradigm}
        \label{fig:subfigure_a}
    \end{subfigure}
    \hfill
    \begin{subfigure}[b]{0.9\linewidth}
        \centering
        \includegraphics[width=\linewidth]{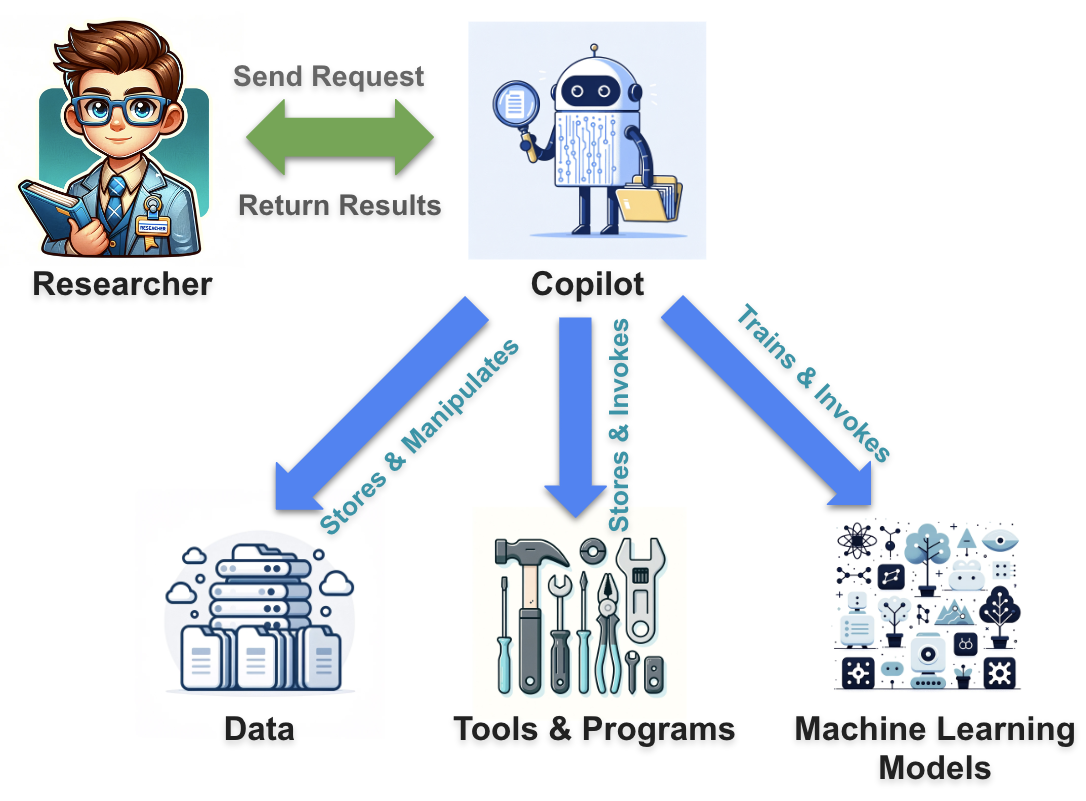}
        \caption{AI-Driven Paradigm}
        \label{fig:subfigure_b}
    \end{subfigure}
    \caption{Shift of Agricultural Data Management Paradigm}
    \label{fig:paradigm_shift}
\end{figure}

In the meanwhile, Large Language Models (LLM) recently start to demonstrate reasoning and tool mastering capabilities. That is, augmented with a set of tools and given a task, LLM need to make plans for the task, select appropriate tools for each step, call the tools and observe the feedback from these tools. Also there can be a observe-and-take-action loop until the task can be accomplished. LLM's reasoning capability makes it feasible to make plans for a task or make decisions about which tool will be used in next step. The application arises naturally where we explore the possibility to fit LLM as the aforementioned orchestrational layer which can coordinate and harness each component of agricultural data management and coordinate each phase of the data processing pipeline. Imagining the tools are not isolated with each other, but form a network where each tool can connect with each other through their inputs and outputs. The tool using intelligence of LLM can probably find new paths in the complicated tool networks and as a result gain new patterns and insight. This will bring us a new data management paradigm in which human effort can be reduced while the outcome of data analysis can be probably maximized. In one word, LLM based copilot will act as a controller in this new data management paradigm, which helps human users to manage data, tools or ML models. Furthermore, as new architectures of LLM-based data management framework being proposed and as new tools, IoT devices or sensors being continuously incorporated, the systems adopting the new paradigm will hopefully grow into a full-fledged autonomous data management system, which can take care of mundane operations with minimal human interference, and help human users to make decisions in data management tasks. Fig \ref{fig:paradigm_shift} illustrates the shift of agricultural data management.

In this paper, we propose and explore the idea of building a LLM-based copilot for autonomous agricultural data management system. We build a proof-of-concept multi-agent system, called ADMA copilot, which can try its best to understand user's intent through interactions in both natural language and GUI, and then make plan for data processing pipeline and accomplish the task automatically. The exiting LLM-based tool using systems generate function calls together with arguments together, probably resulting in hallucinating problem where LLM fabricates the value of arguments without solid ground truth. Different from existing solutions, our system decouple the data flow and control flow, by introducing a meta-program graph which stores all the information about how to use a tool and what a variable stores.  In this way, we solve the problem of hallucinating and increase the controllability and predictability of the copilot. Under the hood, the copilot has three agents collaborating together, one controller which decides the tool called in next step, one input formatter which modifies the value of variables based on user input, and one output formatter which generates UI, based on user's instruction. All the agents base their reasoning on the current status of the meta-program graph, which helps them to make collected decisions. We make comparison with existing agricultural data management platforms by defining a set of criteria to show the superiority and potential of our system in the long run.

We conduct extensive experiments to demonstrate the key features of the our agricultural data management copilot listed as follows:

\begin{itemize}[leftmargin=*]

\item Intelligence: by introducing LLM based agents as program controller, input formatter and output formatter, the copilot demonstrates intelligence in terms of understanding human intent, solution planning and tool/function calling.

\item Autonomy: data management tasks which implicitly contains several steps can be executed automatically with minimal human interference. The design principle of our system is not asking for user input or instruction unless necessary.

\item Efficacy: multi-agent design decouples data flow from control flow, which will assure the predictability of the copilot and eradicate the phenomena of hallucination of LLM.

\item Efficiency: when executing complicated data management tasks, the copilot is way more efficient than human being, thanks to streamlining the whole pipeline.

\item Extensibility: the tool set utilized by the copilot can be extended, by augmenting the meta-program graph, which will guarantee the copilot can have versatile application in the future. 

\item Flexibility: the copilot can understand human input in a flexible way. In some cases, even the user's input is fuzzy and has mistake, the copilot can still get the intent of the user.

\item Privacy: the copilot will protect user's data by asking the user to provide necessary credentials for various services. 

\end{itemize}

Connected to our existing platform ADMA and various external tools, the data management copilot can not only help users to deal with tedious low-level data operations, but also can be evolved to substitute human beings in decision making, which both help to make the shift of agricultural data management paradigm.

The key contributions of our work are summarized as follows:
\begin{itemize}[leftmargin=*]

\item We design and implement Agricultural Data Management and Analytics Copilot (ADMA Copilot), which is a multi-agent copilot for autonomous agricultural data management and analysis. To our knowledge, it is the first of its kind, not only in agricultural data management, but in scientific data management as a whole.
  
 \item As a proof-of-concept product, ADMA Copilot can help to facilitate the shift of agricultural data management paradigm from traditionally human-driven paradigm to AI-driven paradigm. 

\item Extensive demos and evaluations demonstrate the intelligence, autonomy, efficacy, efficiency, extensibility, flexibility and privacy of our system. Comparison is also made between ours and existing systems to show the superiority and potential of our system. 
\end{itemize}

The rest of the paper is organized as follows: Section \ref{sec:related} surveys existing work relevant to our research. Section \ref{sec:framework} presents the design principles and frameworks of our copilot. In Section \ref{sec:evaluation}, we provide the thorough demos and evaluations of the current status of ADMA. Finally, in Section \ref{sec:conclusion}, we conclude the paper by discussing the contributions and suggesting future directions.
\section{Related Work}
\label{sec:related}
\subsection{Tool Learning with Large Language Models}
Tool learning with Large Language Models (LLMs) has emerged recently as a promising paradigm for augmenting the capabilities of LLMs to accomplish complicated tasks, by utilizing LLMs reasoning abilities. Here the tool can be a program, a database, an API endpoint for a cloud based service, a sensor, hardware devices or anything of which the interface can be coded as plan text and thus can be recognized and utilized by LLMs. 

A large corpus of exiting work has been proposed to explore different aspects and phases of Tool learning with LLMs \cite{qu2024toolsurvey,shen2024llm,mialon2023augmented}. Tool learning with LLMs is in fact an umbrella concept which encompasses a lot of subcategories, including augmenting LLMs with external knowledge \cite{nakano2021webgpt,zhuang2023toolqa,lewis2020retrieval,fan2024survey}, augmenting LLMs with mathematical reasoning ability \cite{he2023solving,gou2023tora}, augmenting LLMs with program executors \cite{gao2023pal,wang2024executable}, and augmenting LLMs with domain specific tool sets \cite {bran2023chemcrow,zhang2024multimodal}. Generally speaking, there are four phases in tool learning with LLMs: task planning, tool selection, tool calling and response generation\cite{qu2024toolsurvey}. In task planning, a task is decomposed into several sub-tasks, by utilizing LLM's reasoning capabilities \cite{wei2022chain,shen2024hugginggpt,shi2024chain,zhuang2023toolchain,ge2024openagi,qin2023toolllm}. In tool selection, the most appropriate tool is selected by LLM for each phase generated from task planning \cite{qiao2023making,liu2024toolnet}. In tool calling phase, the selected tool is called, based on given interface of the tool. Again LLM will be used to format the calling interface, that is, configure the order and value of arguments for a method call \cite{song2023restgpt,yang2024gpt4tools,li2023tool}. Nowadays, large language models such as GPT-4 can output function calling, given the description of the tool and the user input \cite{openai_function_calling_2023}. Lastly, the output of the tools can be combined and formatted for output, by utilizing LLM's language organization capability \cite{schick2024toolformer}. Each phase can utilize LLM and thus there will be several agents collaborating together for the whole process. 

The existing works explore different frameworks to implement tool learning with LLMs, however, almost all the works don't decouple data flow from control flow, which may result in hallucination problem and in turn fail a task. In our work, we generate a meta-program graph, as a representation of the tools and their affiliated data and use three agents: controller, input formatter and output formatter to deal with control flow and data flow respectively. In this way, we can increase the predictability of the result generated by the LLM-based agents and thus increase the success rate of the task solving process.

\subsection{Agricultural Data Management Systems}

The digital revolution in agriculture has paved the way for advanced data management platforms specifically designed to tackle the unique challenges and demands of agricultural research and production. Several notable platforms have emerged in related sectors.

CyVerse \cite{merchant2016iplant}, initially developed for plant genomics, has expanded into a comprehensive platform, offering life scientists robust computational infrastructure to manage large datasets and conduct complex analyses. As a cloud-based solution, CyVerse supports collaborative efforts, enabling seamless sharing of data and tools. Its focus on scalability and interoperability has made it a crucial tool for modern agricultural data management. GARDIAN \cite{jouini2021evaluation}, the CGIAR’s (Consultative Group for International Agricultural Research) data discovery platform, allows users to find research datasets and publications from various CGIAR Centers and Programs. By aggregating data from multiple agricultural research projects, GARDIAN offers a holistic view of global agricultural research, fostering interdisciplinary collaboration. GEMS \cite{senay2022big} integrates genomic, phenotypic, and environmental data, offering tools for analyzing and visualizing large-scale agricultural datasets, thereby linking genomics with environmental factors in agricultural research. TERRAREF \cite{lebauer2020terra} focuses on delivering high-resolution sensor data for plant research, using tools such as cameras and drones to gather detailed information on plant growth, health, and environmental conditions. This extensive dataset aids researchers in exploring the intricate relationships between plants and their environments. The evolution of Smart Farming into Agriculture 5.0 \cite{saiz2020smart} emphasizes the critical role of data in optimizing farm management for sustainability and economic efficiency. Lastly, ADMA \cite{pan2023transforming} introduces an innovative agricultural data management system designed according to FAIR principles, incorporating features such as intelligence, interactivity, scalability, flexibility, open-source pipeline management, and enhanced privacy and security.

Several platforms and studies emphasize the importance of integrated data management systems for addressing specific agricultural topics or utilizing advanced technologies. For instance, Agricultural Remote Sensing Big Data \cite{huang2018agricultural} explores the management and application of remote sensing big data in agriculture, introducing a four-layer-twelve-level (FLTL) framework designed to handle remote sensing data for precision farming and local farm analyses. Data Warehouse and Decision Support on Integrated Crop Big Data \cite{ngo2020data} presents a thorough analysis of data warehousing and decision support systems, demonstrating their capability in leveraging integrated crop big data through the development of a continental-level Agricultural Data Warehouse (ADW). Big Data and Machine Learning (ML) With Hyperspectral Information in Agriculture \cite{ang2021big} reviews critical research utilizing big data, machine learning, and deep learning, with a specific focus on processing hyperspectral and multispectral agricultural data. It highlights the potential of ensemble machine learning and scalable parallel discriminant analysis in managing agricultural big data. Lastly, Blockchain for Sustainable e-Agriculture \cite{dey2021blockchain} discusses the intersection of blockchain technology and IoT in e-agriculture, emphasizing how blockchain can enhance agricultural value chains, strengthen IoT networks, and ensure data validation, security, and privacy.

\section{Framework}
\label{sec:framework}
To support the FAIR principles and the proposed key features such as intelligence, autonomy, efficacy, efficiency, extensibility, flexibility and privacy, we design and implement a proof-of-concept system called Agricultural Data Management and Analytics Copilot (ADMA Copilot).

\subsection{Shift of Paradigm}
\label{subsec: shift_of_paradigm}
Thanks to the emerging reasoning and tool learning ability of large language models, there will be a shift of agricultural data management paradigm. In the traditional paradigm, researchers need to understand every details of the data and tools (or ML models) at their hands, such as location, semantics and format of data, and the location, semantics and the interface of the tools (or ML models). When conducting data management, researchers need to manipulate the data, tools or ML models in person. So we can call the traditional paradigm as user-driven paradigm. On the other hand, in the new data management paradigm, an LLM-based copilot will take over to manipulate the data management pipeline, and the researchers will only send high-level requests to the copilot to initiate the pipeline and get involved when required. We can call the new paradigm as AI-driven paradigm. Fig \ref{fig:paradigm_shift} illustrate the shift of agricultural data management shift.

\subsection{System Components}
\begin{figure}[htbp]
\centerline{\includegraphics[width=1\linewidth]{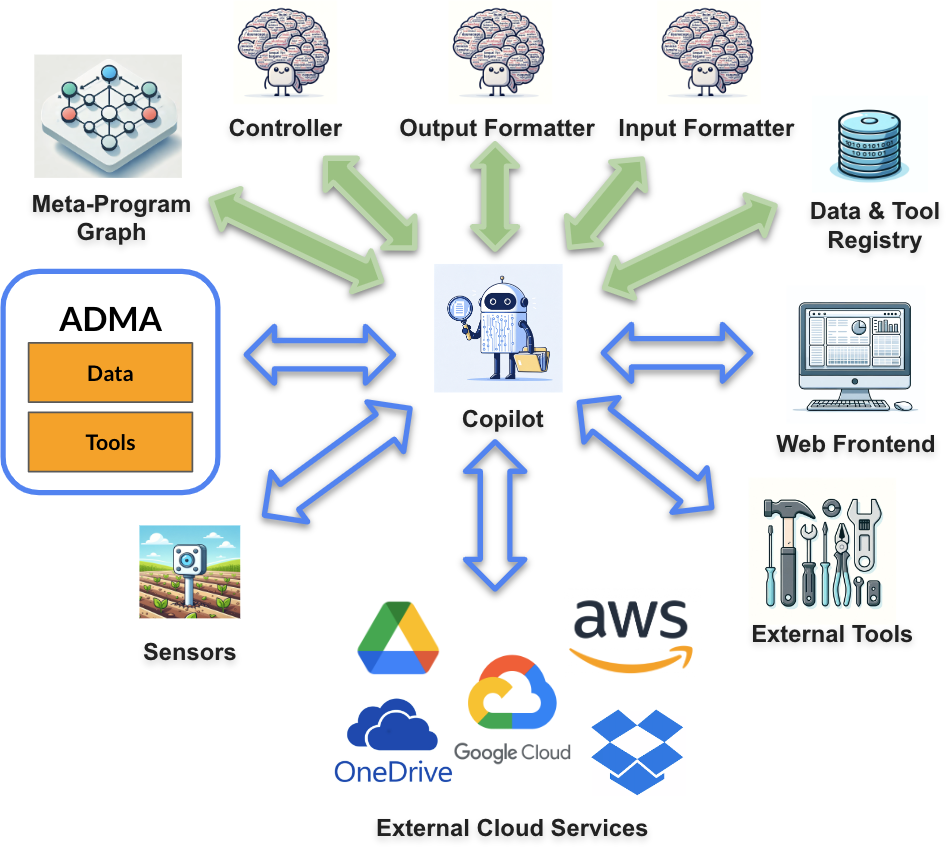}}
\caption{Main Components of ADMA Copilot}
\label{fig:framework}
\end{figure}

There are two types of components in our framework: internal components and external components. Internal components are the integral parts of the ADMA Copilot, including  a copilot server, three LLM-based Agents, a meta-program graph, a data\&tool registry. The ADMA Copilot talks with external components such as the original ADMA system, sensors, external cloud storage and computing services, external tools \& APIs, and finally, a web front-end. Figure~\ref{fig:framework} illustrates the the main components ADMA Copilot and their interactions.  In The following subsections, we'll introduce each main components.

\subsubsection{Copilot Server}

The copilot sever is the hub of the whole system, which is responsible for coordinating other components. The copilot server receives instructions from the web front-end and output the results to the front-end. When receiving a task, the copilot server will search in its data \& tool registry for relevant tools to use and build a meta-program graph. The copilot server also talks to three LLM-based agents and meta-program graph, to decide the tools to call in each step and to format the input/output of the meta-program. When the tool to call is determined, the copilot server will reach out to the specified tool, execute the call, and retrieve the results from the tool. 

The design principle of the copilot server is to decouple the programmatic part from the "intelligent" part, by only keeping the programmatic part within its own logic and off-loading the intelligent part to the three LLM-based agents. In this way, the logic of copilot server can be kept at its minimal.

\subsubsection{LLM-based Agents}
There are three LLM-based agents: a program controller, an input formatter and an output formmater, all of which will accept current task, meta-program graph and the execution history as input. The program controller is responsible for determining the control flow of the program, namely, which tool to execute in next step, by looking into the status of current meta-program graph and taking into account of the task to accomplish. The program controller has the highest requirements for the reasoning capability of the underlying LLM, because it needs to analyze current task and conduct the planning accordingly, and for some complicated tasks, the planning always requires high reasoning capability.The other two LLM-based agents, input formatter and output formmater, control the data flow, by translating the input to the value of the variable or translating the value to output, respectively. 

Three agents are coordinated by the copilot server and are triggered when required to keep the control flow and data flow of the program on track, thus finally solve the task proposed by the end user.

\subsubsection{Meta-Program Graph}
Meta-program graph is a indispensable component which encodes the meta-structure of the tools and data. Through meta-program, the for-mentioned LLM-based agents can understand what a tool or data is about and make collected decisions on controlling the control flow and data flow in completing the task. It is the meta-program that makes data flow decoupled from the control flow and as a result, increase the predictability of the result.

\begin{figure}[htbp]
\centerline{\includegraphics[width=1.01\linewidth]{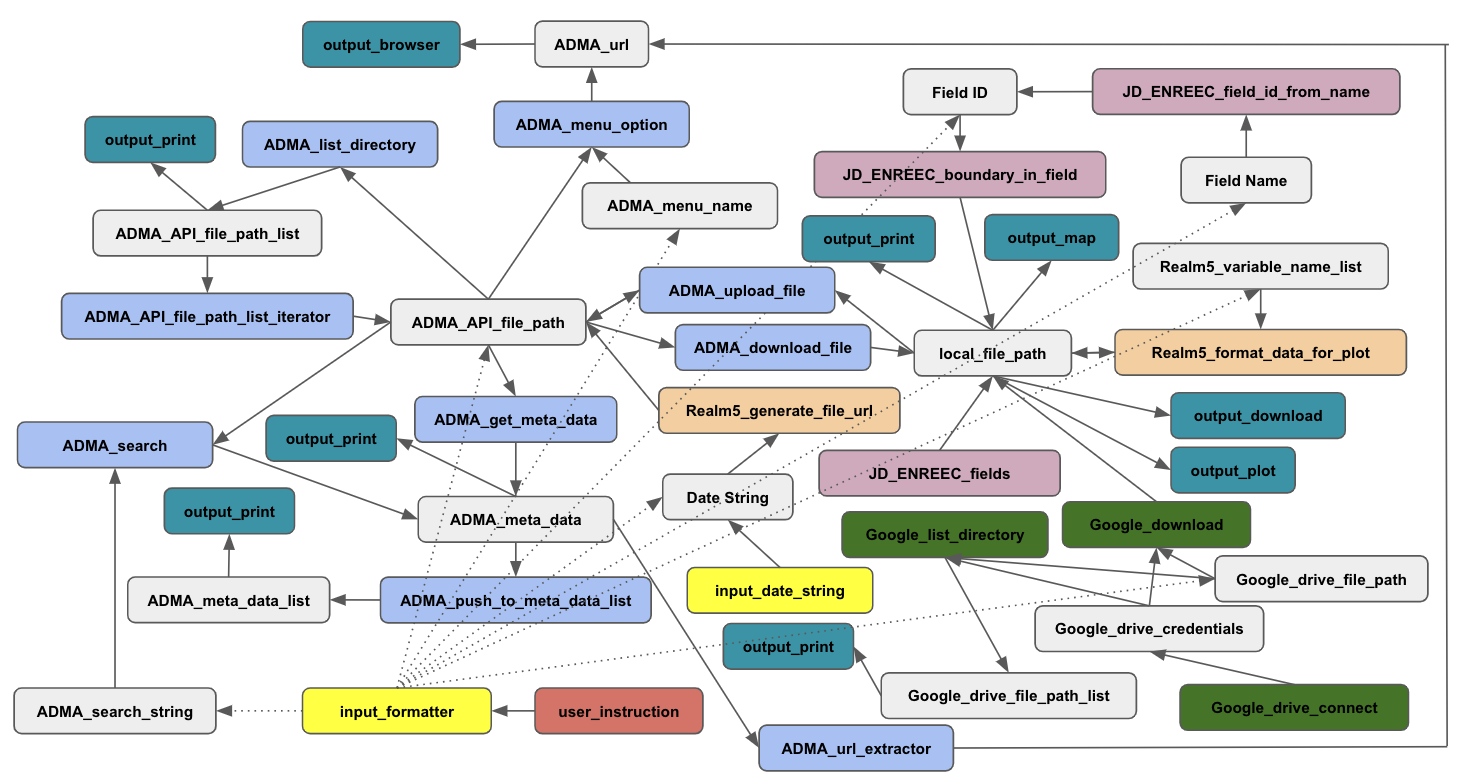}}
\caption{An Example of Meta-Program Graph}
\label{fig:meta_program_graph}
\end{figure}

Figure \ref{fig:meta_program_graph} illustrates an example of meta-program graph, in which the boxes in color represent various types of tools and the boxes in gray represent the variables, the arrows between each pair of boxes indicates the direction of data flow. Each task will begin from the user input (red box). After translated by input formmater, the values contained in the instruction will start to flow on the meta-program graph. In each step, the controller will be called to decide the routing on meta-program graph. If an output condition is triggered and the output formatter will be called to format a result, in natural language, or in UI. For instance, if the user asks for some knowledge, the output will normally be in natural language, if the user asks for a data visualization, then a plot will be displayed, or if the user asks for data downloading, a download button will pop up. 

We design the meta-program in a concise format, which contains all the necessary information about the data and tools, while keeping it small enough to fit in the context window of a typical LLM. Also the meta-program graph is extensible to contain hundreds or thousands of tools, which will help to solve complicated tasks containing tens or hundreds of steps of actions. 

\subsubsection{Data \& Tool Registry}
The meta data about tools and data can be store in data \& tool registry for the copilot to reference. When the number of tools grows really large, it is not possible to store the information about all the tools and affiliate variables in meta-program graph, so in this case, copilot will first retrieve relevant tools to construct the meta-program graph, and also retrieve the data to initialize relevant variables. Also the registry can store user and session information, which makes it act as a long-term memory of the copilot.

\subsubsection{ADMA}
\label{subsubsec: data_storage}
ADMA is short for Agricultural Data Management and Analytics, which is the system we developed before \cite{pan2023transforming}. Here we incorporate ADMA into the framework by connecting the copilot with ADMA. Through ADMA's API, the copilot can manipulate the data and the tools hosted on ADMA, just as a human user. In the copilot's perspective, ADMA can be considered as a set of external tools, which can be integrated into the data management pipeline. We'll demonstrate some use case of the copilot involving ADMA in the following section.

\subsubsection{Sensors \& IoT Devices}
ADMA Copilot can be connected to various sensors and IoT devices on the field, through APIs or specifically designed interface. The copilot will treat the sensors the same way as other tools, to collect on-field information such as air temperature, humidity, precipitation, wind speed, soil moisture, nitrogen, carbon dioxide, fertilizer application and etc, which is extremely important for certain tasks involving decision making.  Here in our proof-of-concept system, we connect the copilot with John Deere \cite{deere2024} API and Realm5 \cite{realmfive} API, to collect field operation information and weather data respectively. We'll demonstrate some use case of the copilot calling sensors through these APIs in the following section.

\subsubsection{External Cloud Services}
ADMA copilot can make full use of existing cloud storage and computing service, such as One Drive, Google Drive, Google Cloud, Amazon AWS or Dropbox. These cloud services provide more options for the user to store their data, run their program or host the ML/AI models. Copilot will register the interface for these cloud services and help the user to manage their data, tools or models on these service. For our proof-or-concept copilot, we connect it with Google Drive. Autonomous data movement between Google Drive and ADMA is demonstrated in the next section.

\subsubsection{External Tools}
Copilot can be connected to various other external tools, through APIs or specifically designed interface. When new tools get connected, they will be registered in the data \& tool registry and later when needed, copilot can discover and call them. These feature assures the extensibility of our system and make it applicable to any application scenario.

\subsubsection{Web Front-end}
A conversation based web front-end is provided to the end user, who can type in any data management instruction to the copilot. Once the copilot completes the instruction (or fails it), the final response will be displayed back to the end user on the web front. The specific format of the response will depend on the instruction. It can be in natural language, a table, a plot, a UI gadget, a web page or a image. The following section will contain several use cases about the results showing on the web front-end.

\section{Evaluation}
\label{sec:evaluation}
In this section, we demonstrate the use cases of ADMA Copilot to prove its core features such as intelligence, autonomy, efficacy, efficiency, extensibility, flexibility and privacy.

\subsection{Interface}
\begin{figure}[htbp]
\centerline{\includegraphics[width=1\linewidth]{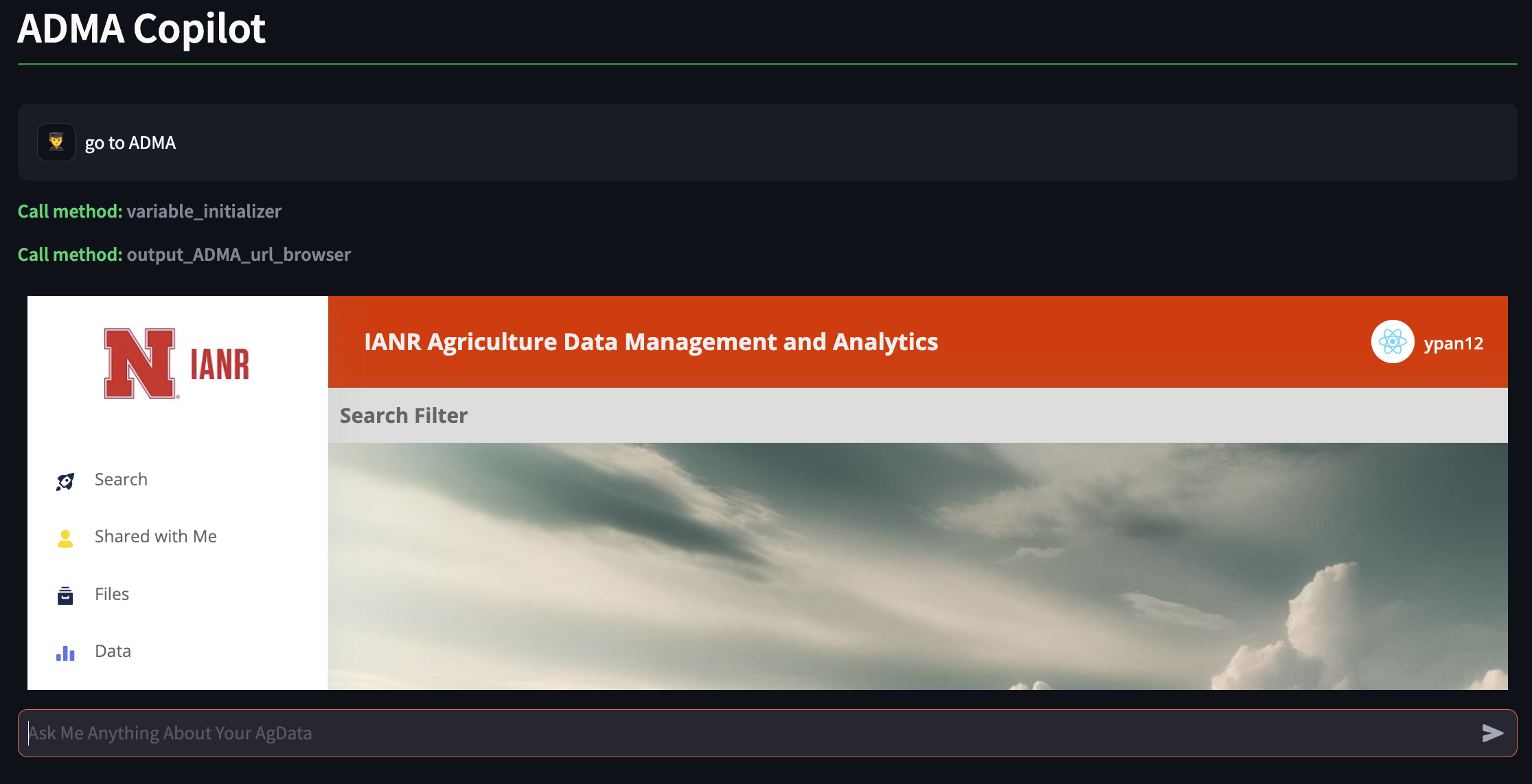}}
\caption{A Demo for Conversational UI of ADMA Copilot}
\label{fig:demo_interface}
\end{figure}

ADMA Copilot has a conversational interface, where the user can type in the instruction into the chat box at the bottom of the web page. Then the results will be displayed above the chat box. All the chat history within a session will be preserved. In Figure \ref{fig:demo_interface}, when the user asks "go to ADMA", then the main page of ADMA will be displayed.  ADMA Copilot will also display the steps it has gone through.

\subsection{Intelligence}
\begin{figure}[htbp]
\centerline{\includegraphics[width=1\linewidth]{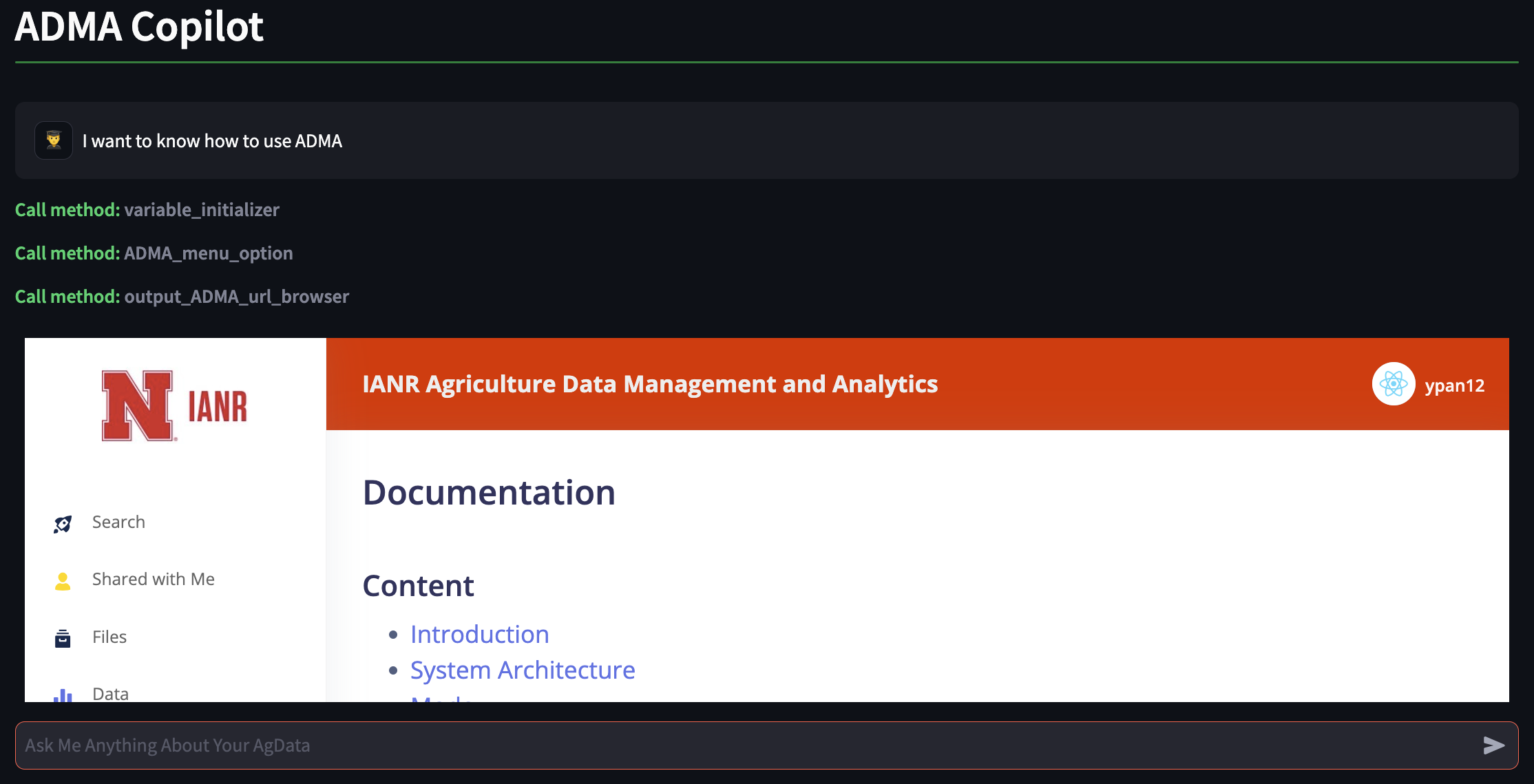}}
\caption{A Demo for User Intent Understanding}
\label{fig:demo_intelligence}
\end{figure}

ADMA Copilot can understand user intent in natural language. Sometimes, even the user input a fuzzy or somewhat ambigous instruction, the copilot will try its best to interpret the user's intent. Figure \ref{fig:demo_intelligence} illustrate a case in which the user type in the instruction: "I want to know how to use ADMA." Then the copilot will display the documentation page of the ADMA because this is the result which is most appropriate in this context.

\subsection{Autonomy}
One of the key features of ADMA copilot is autonomy. It is designed to try its best to accomplish the tasks proposed by users autonomously, except extra user input is required. For each task, the program controller of the copilot will route on the meta-program graph and complete several steps until the task is finally accomplished.

\begin{figure}[!t]
    \centering
    \begin{subfigure}{1\linewidth}
        \centering
        \includegraphics[width=\linewidth]{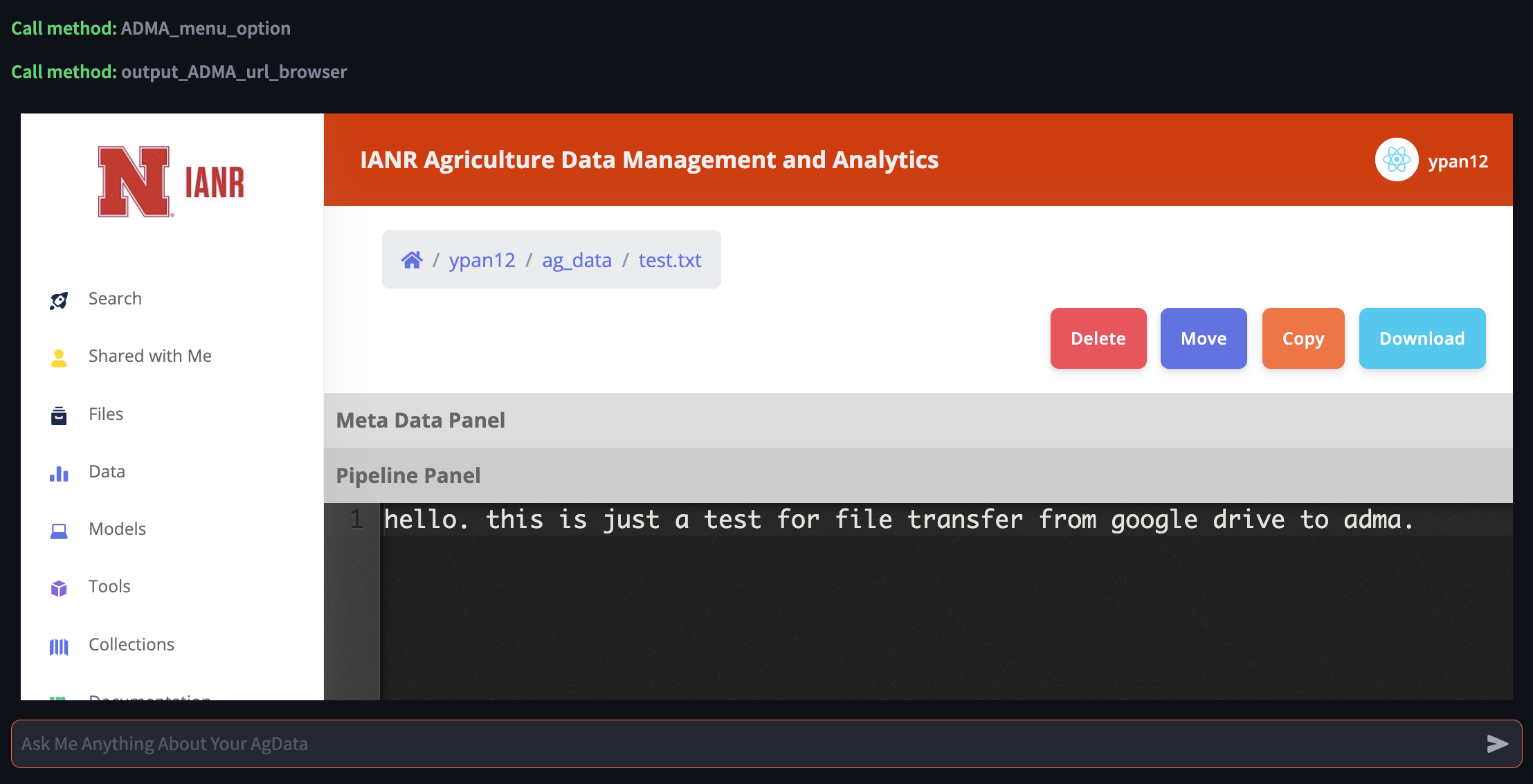}
        \caption{Transfer File from Google Drive to ADMA and Open the Page }
        \label{fig:demo_transfer}
    \end{subfigure}%
    \hfill
    \begin{subfigure}{1\linewidth}
        \centering
        \includegraphics[width=\linewidth]{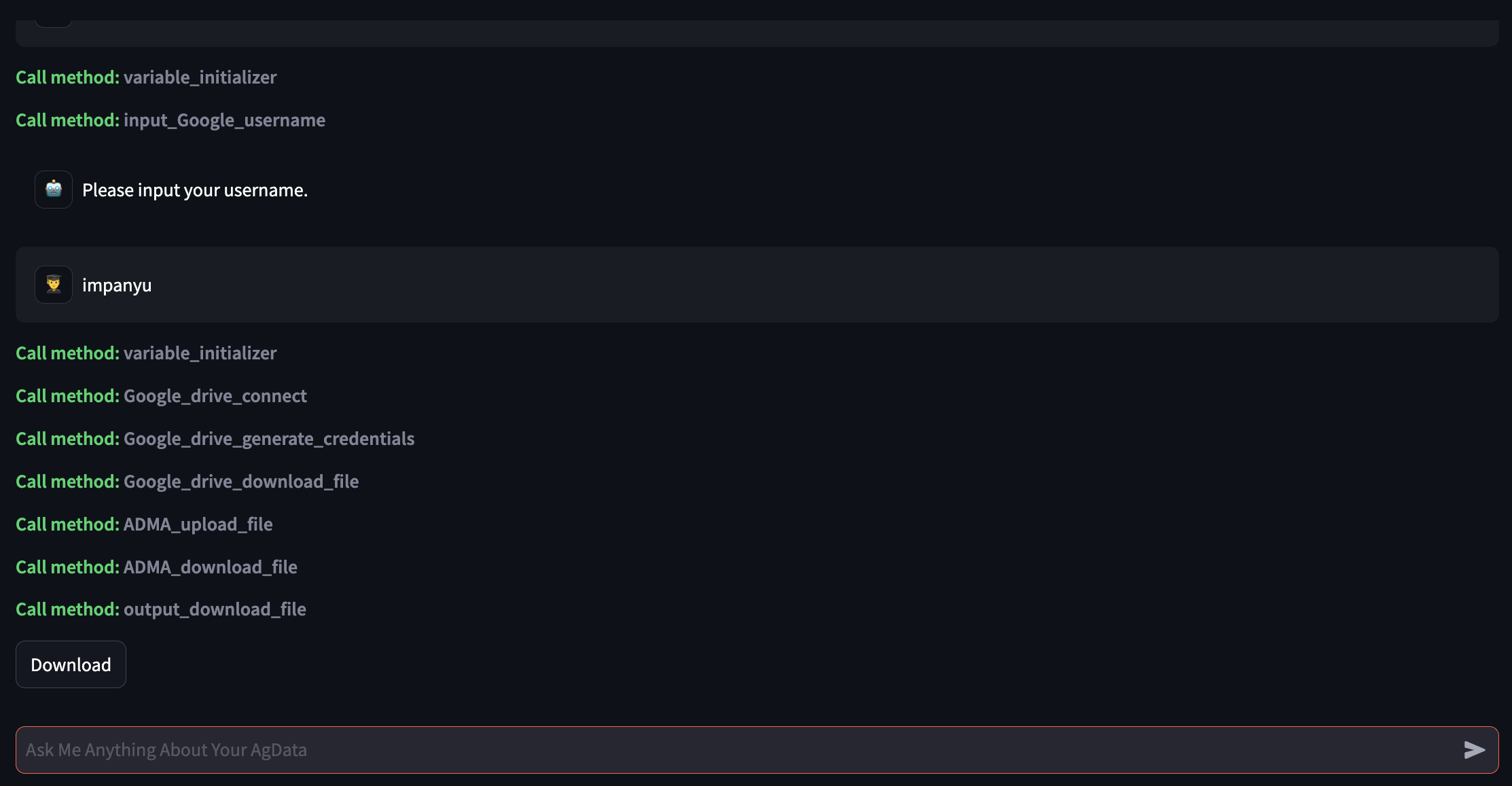}
        \caption{Transfer File from Google Drive to ADMA and then Download}
        \label{fig:demo_transfer_and_download}
    \end{subfigure}%
    \hfill
    \begin{subfigure}{1\linewidth}
        \centering
        \includegraphics[width=\linewidth]{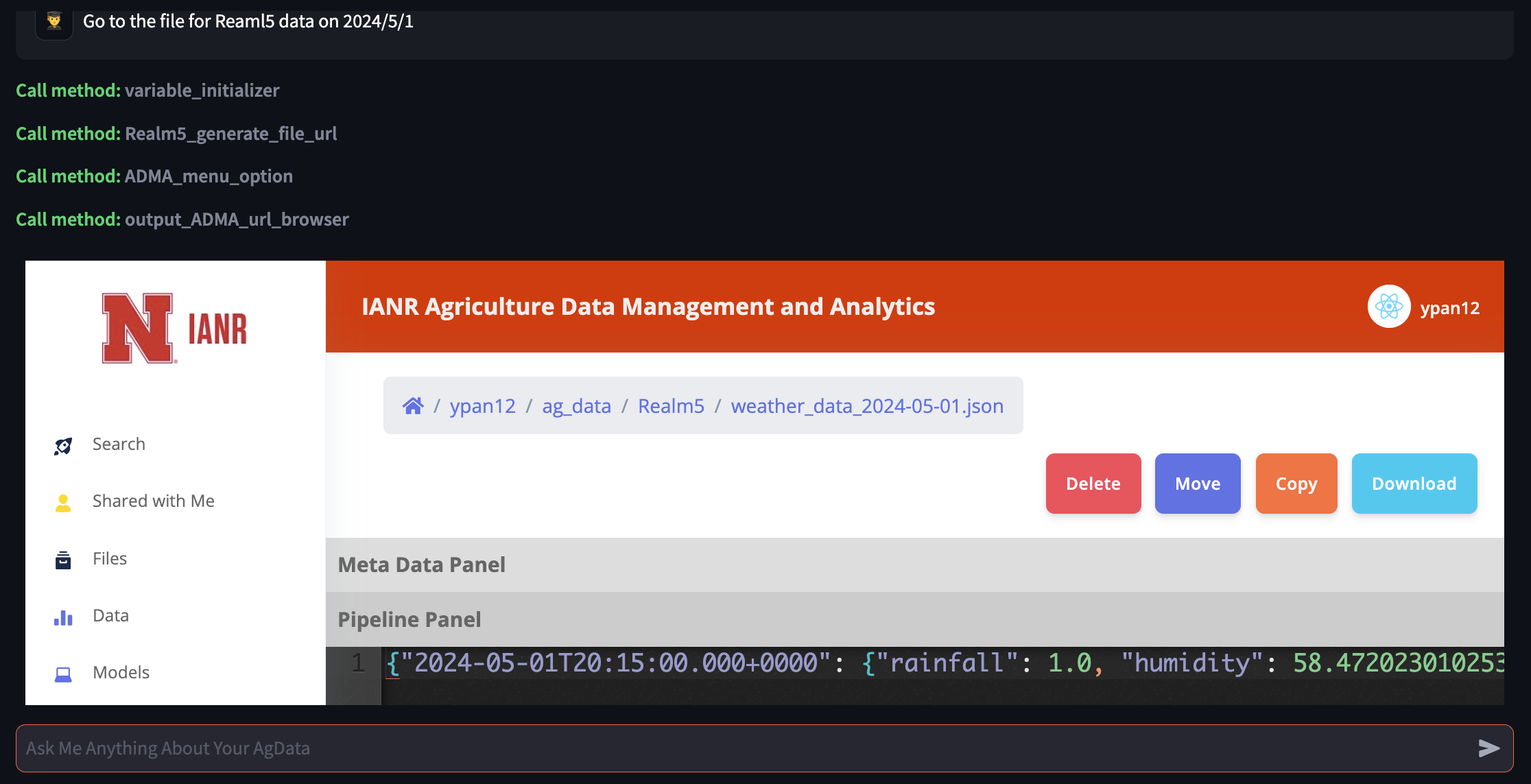}
        \caption{Open the file for Realm5 Data on 2024/5/1}
        \label{fig:demo_realm5_page}
    \end{subfigure}%
    \hfill
     \begin{subfigure}{1\linewidth}
        \centering
        \includegraphics[width=\linewidth]{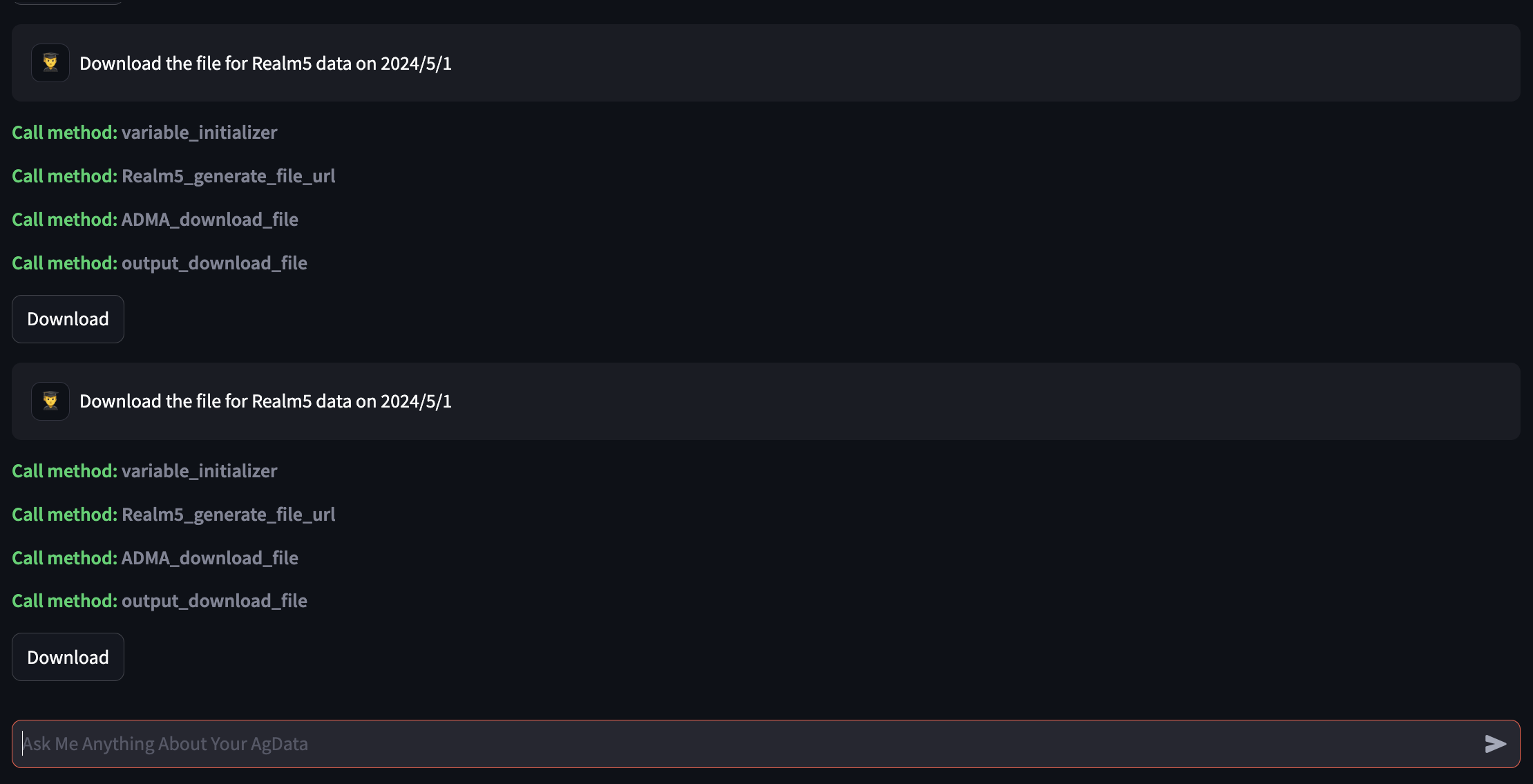}
        \caption{Download the file for Realm5 Data on 2024/5/1}
        \label{fig:demo_realm5_download}
    \end{subfigure}%
    \hfill
   
    \caption{Demo for Autonomous Task Completion}
    \label{fig:demo_autonomy}
\end{figure}

Figure \ref{fig:demo_autonomy} demonstrates some use cases of autonomous task completion. In Figure \ref{fig:demo_transfer}, the user type in  "Transfer the adma\_test/test.txt on my google drive to my ADMA root folder, and open the uploaded file". The copilot decomposes the task into several steps, such as call the input formatter, authenticate the user for Google drive, download the file from Google Drive, upload the file to ADMA, then open the page of the uploaded file.  In Figure \ref{fig:demo_transfer_and_download}, the user type in "Transfer the adma\_test/test.txt on my google drive to my ADMA root folder, and then download the uploaded file", similarly the copilot accomplish the task by first downloading the file from Google drive and then uploading it to ADMA, followed by poping up a download button for the user to click. In Figure \ref{fig:demo_realm5_page}, the user types in "go to the file for Realm5 data on 2024/5/1", again the copilot decompose the task into several steps: initializing the date and menu name, generating the path for the file on ADMA, generating the url for the file on ADMA, displaying the web page for the file. In Figure \ref{fig:demo_realm5_download}, the user asks "download the file for Realm5 data on 2024/5/1", the copilot first generates the url for the file on ADMA as previous task, and then call download API of ADMA to download the file, followed by popping up a download button.

Through these use cases, we can see ADMA Copilot has the basic capability to understand user's intent and complete tasks automatically, with minimal user interference.

\subsection{Efficacy}
The design of our copilot reduces the phenomena of hallucination of LLM to its minimal, by decoupling data flow from control flow. That is, when generating the calling interface of any tool, instead of asking LLM-based controller to do so, we use the value of the variable within meta-program graph. If there are not enough information, the copilot will ask for further input from the user, rather than fabricate some plausible value. This will make sure the predictability of the final results.

\begin{figure}[!t]
    \centering
    \begin{subfigure}{1\linewidth}
        \centering
        \includegraphics[width=\linewidth]{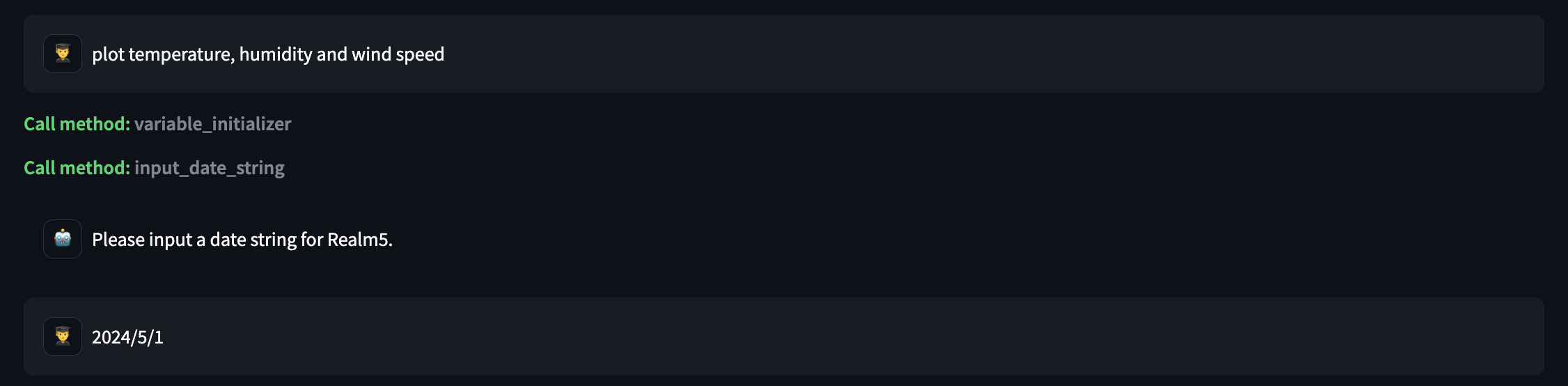}
        \caption{Ask for User Input when Unsure}
        \label{fig:demo_efficacy_1}
    \end{subfigure}%
    \hfill
    \begin{subfigure}{1\linewidth}
        \centering
        \includegraphics[width=\linewidth]{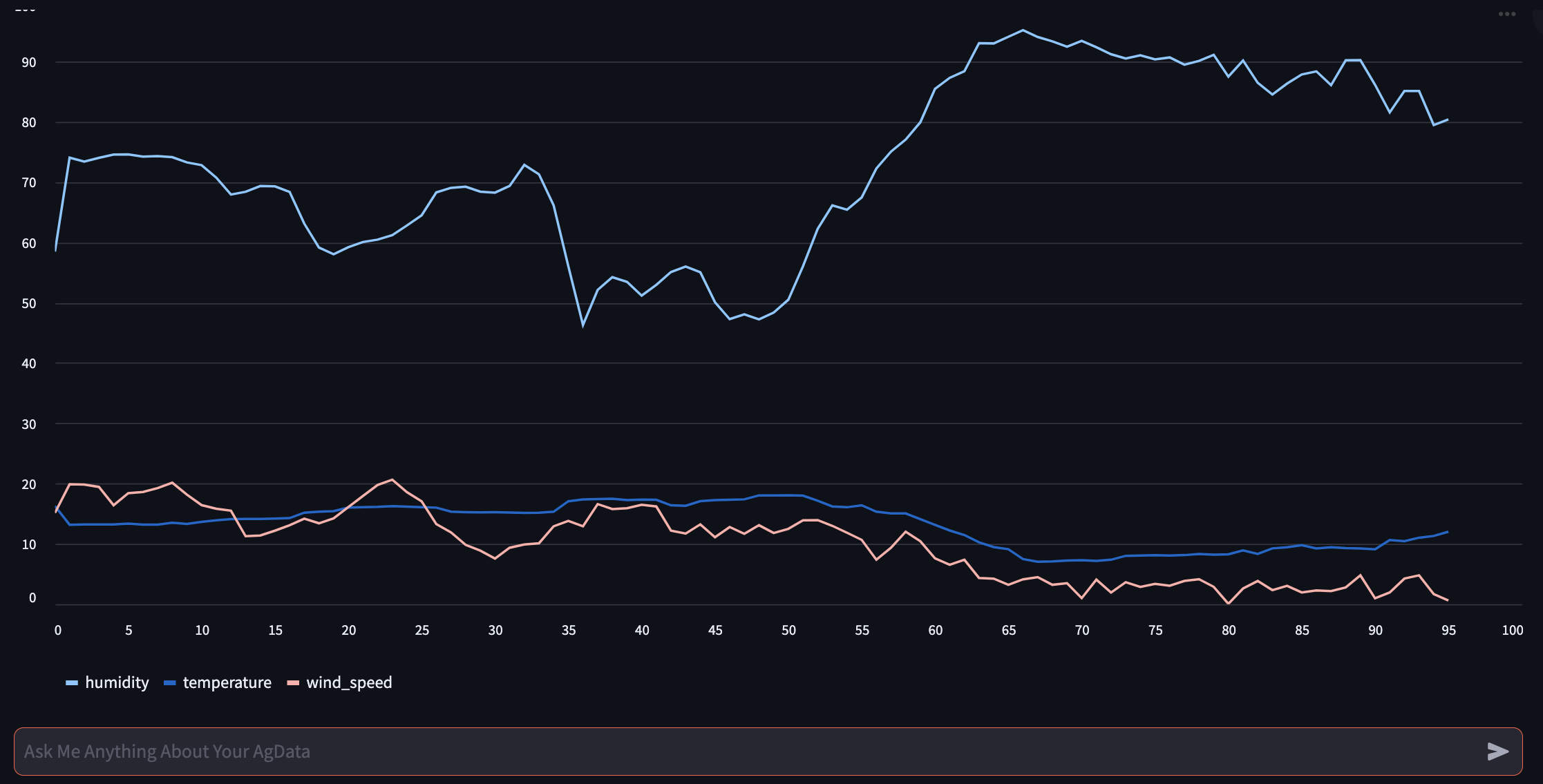}
        \caption{Display the Result after Further Input}
        \label{fig:demo_efficacy_2}
    \end{subfigure}%
    \hfill

    \caption{Demo for Efficacy}
    \label{fig:demo_efficacy}
\end{figure}

Figure \ref{fig:demo_efficacy} illustrates a case when the user inputs "plot temperature, humidity and wind speed" in \ref{fig:demo_efficacy_1}, since the copilot is not sure the time span of the data to draw, it will ask for further input: "Please input a data string for Realm5.". After the user provide the date "2024/5/1", the copilot will retrieve the relevant data and plot it to the user as in \ref{fig:demo_efficacy_2}.

\subsection{Efficiency}
Autonomy not only brings about convenience, but also efficiency. For some complicated tasks which contain several steps, the copilot demonstrates its efficiency compared with human users. Here we conduct a comparison between ADMA copilot and human user, in terms of task completion time for several data management tasks. 

\begin{figure}[htbp]
\centerline{\includegraphics[width=1\linewidth]{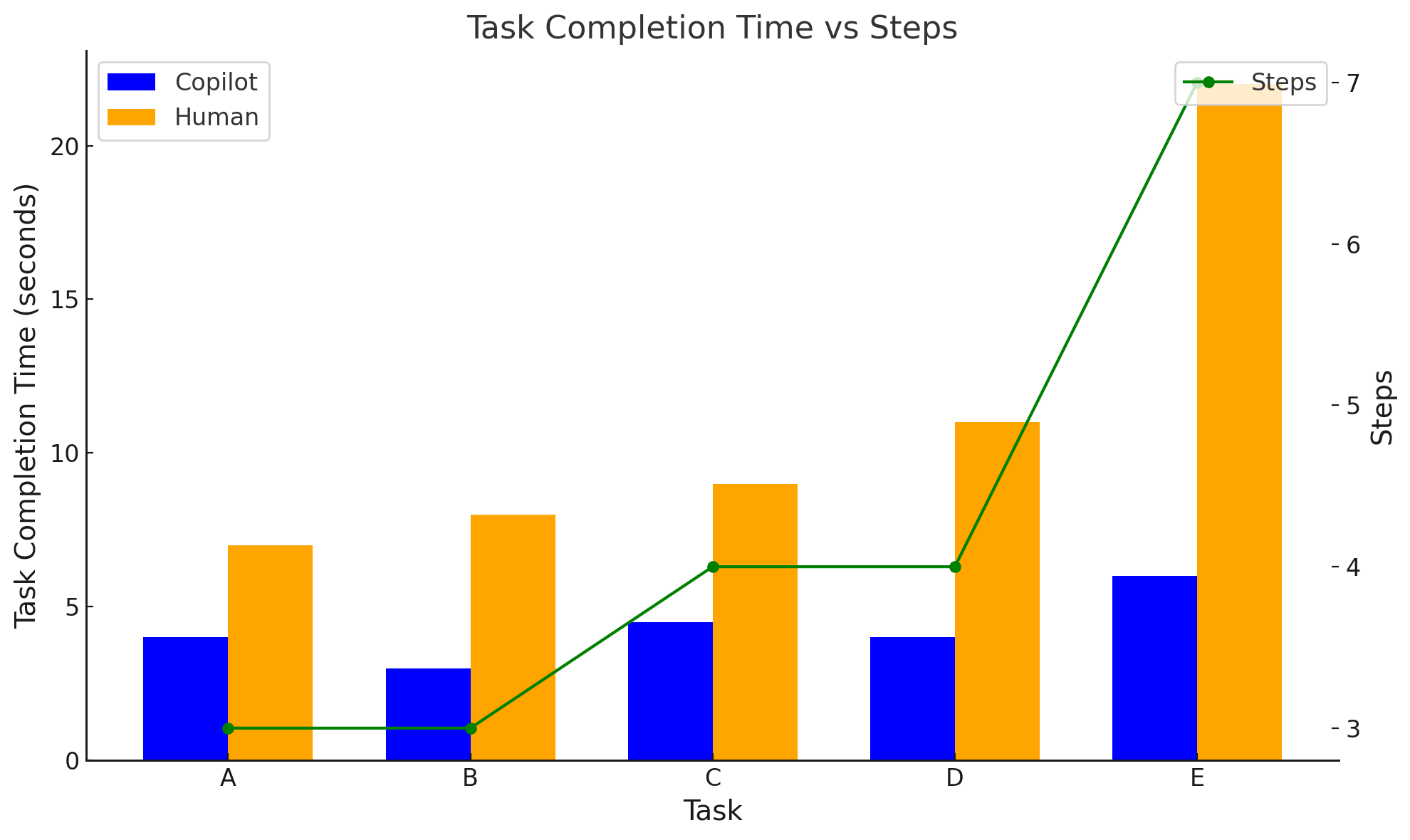}}
\caption{Comparison of Task Completion Time between Copilot and Human Users.}
\label{fig:demo_efficiency}
\end{figure}

In Figure \ref{fig:demo_efficiency}, we choose 5 tasks as follows:
\begin{enumerate}[label=\Alph*.]
\item  "Go to directory 1 under root folder of ADMA", 
\item "Check the meta data of calculate\_ndvi.py on ADMA", 
\item  "Open the page on ADAM, the name of which contains the keyword soil", 
\item  "Download the file for Realm5 data on 2024/5/1", 
\item  "Transfer the adma\_test/test.txt on my google drive to my ADMA root folder, and open the uploaded file"
\end{enumerate}

The bars represent the average task completion time for copilot and human user, for each task. The green curve represents the number of steps the copilot has gone through for each task, which provides some clue about the complex level for a task. We can see for more complicated tasks, both the copilot and human user take longer time, however, the copilot outperforms human user in all tasks, and also the copilot keeps the completion time almost constant, regardless of the increasing complex level of tasks.

\subsection{Extensibility}
\begin{figure}[htbp]
\centerline{\includegraphics[width=1\linewidth]{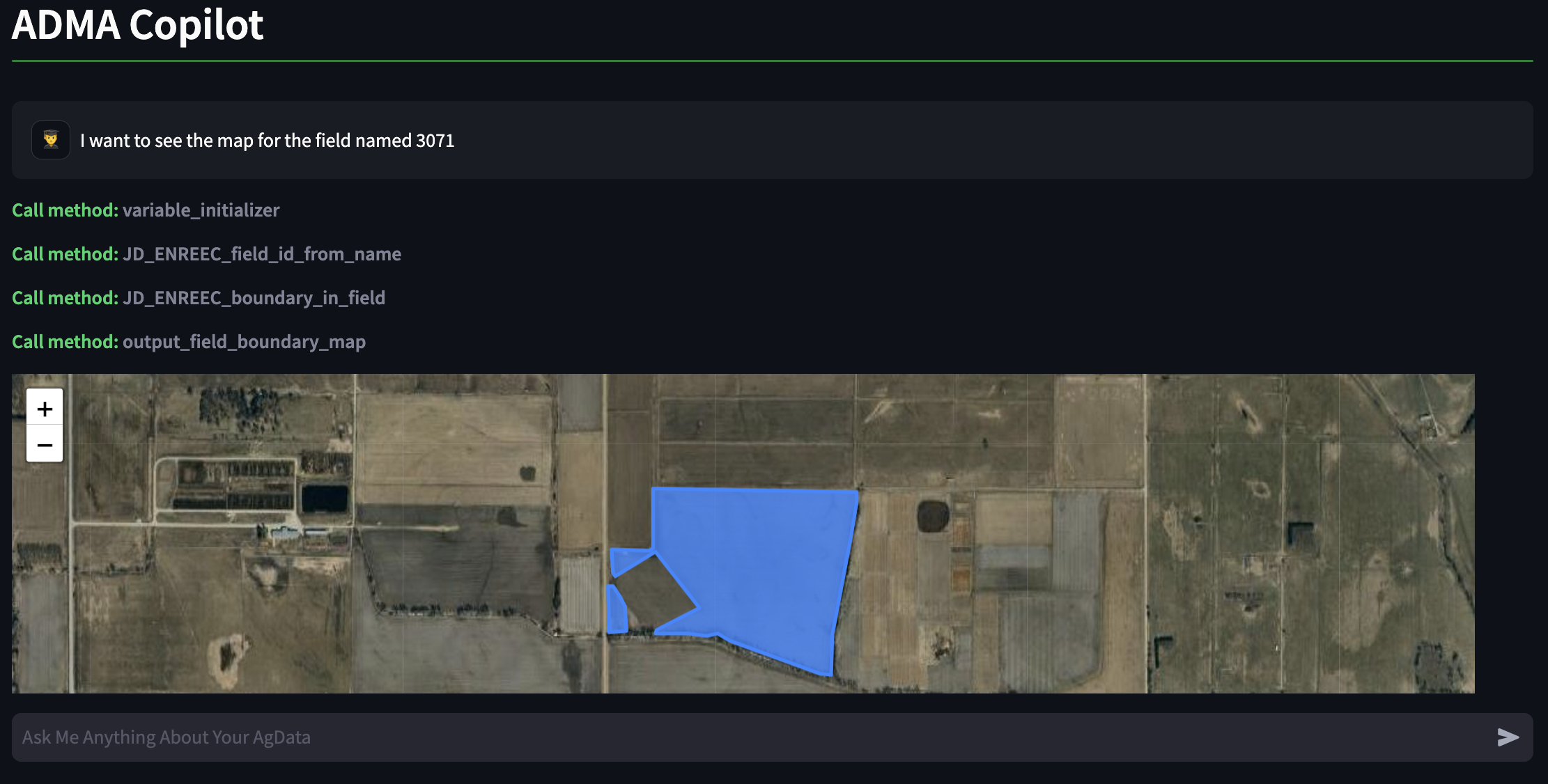}}
\caption{Demo of Drawing a Map for a Field after Plugging in the Map UI}
\label{fig:demo_extensibility}
\end{figure}

To accommodate more tools, ADMA Copilot is designed to be able to incorporate new tools which will be incorporated in the future. Given a documentation of any new API or function interface, the only thing we need to do is to plug in the tool into the meta-program graph and inform the copilot how to use the new tools. Then all things left will be handled to the copilot without any changes as before. The copilot will make calls to the new tools when required. For instance, when we design a new user interface which can draw a map for the fields on our extension farm, we can just plug in the description of the UI tool into meta-program graph, and then when the user inputs "I want to see the map for the field named 1863N", the output will be like in Figure \ref{fig:demo_extensibility}

\subsection{Flexibility}

Thanks to the reasoning capability of the underlying LLM, our copilot demonstrates some degree of flexibility in terms of understanding the user's intent based on a wide spectrum of user input. The user can be free from precisely phrasing the instruction, and the user input can be fuzzy or even ambiguous. Nevertheless, the copilot will try its best to interpret the user's instruction and execute the task accordingly.

\begin{figure}[!t]
    \centering
    \begin{subfigure}{1\linewidth}
        \centering
        \includegraphics[width=\linewidth]{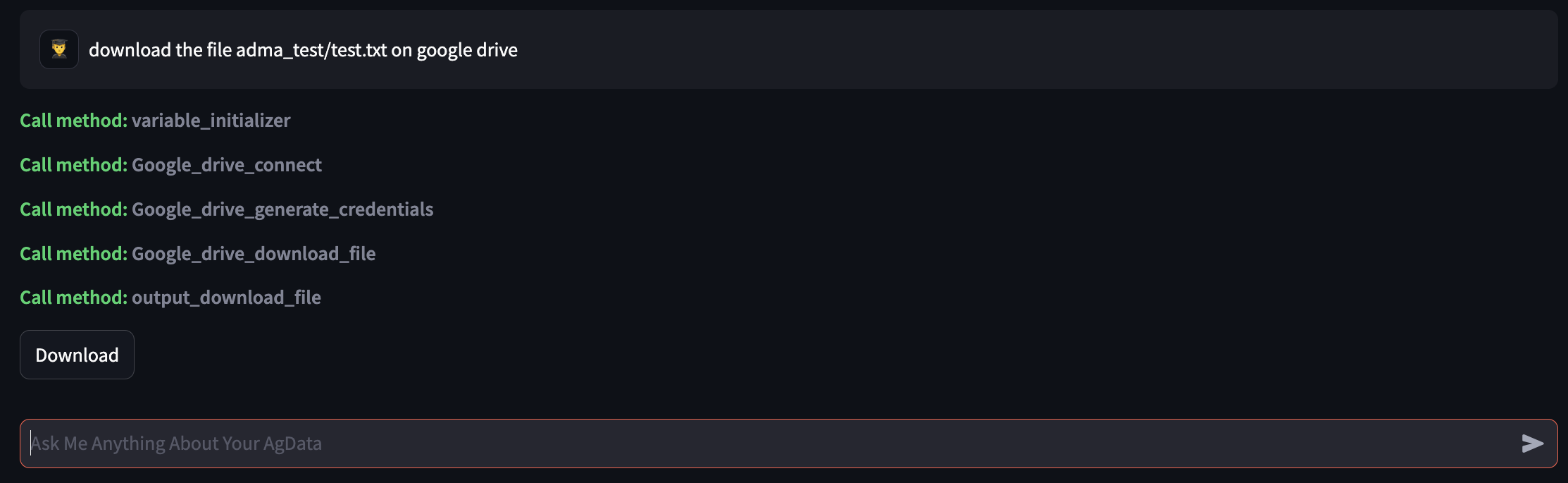}
        \caption{Download A File from Google Drive}
        \label{fig:demo_flexibility_1}
    \end{subfigure}%
    \hfill
    \begin{subfigure}{1\linewidth}
        \centering
        \includegraphics[width=\linewidth]{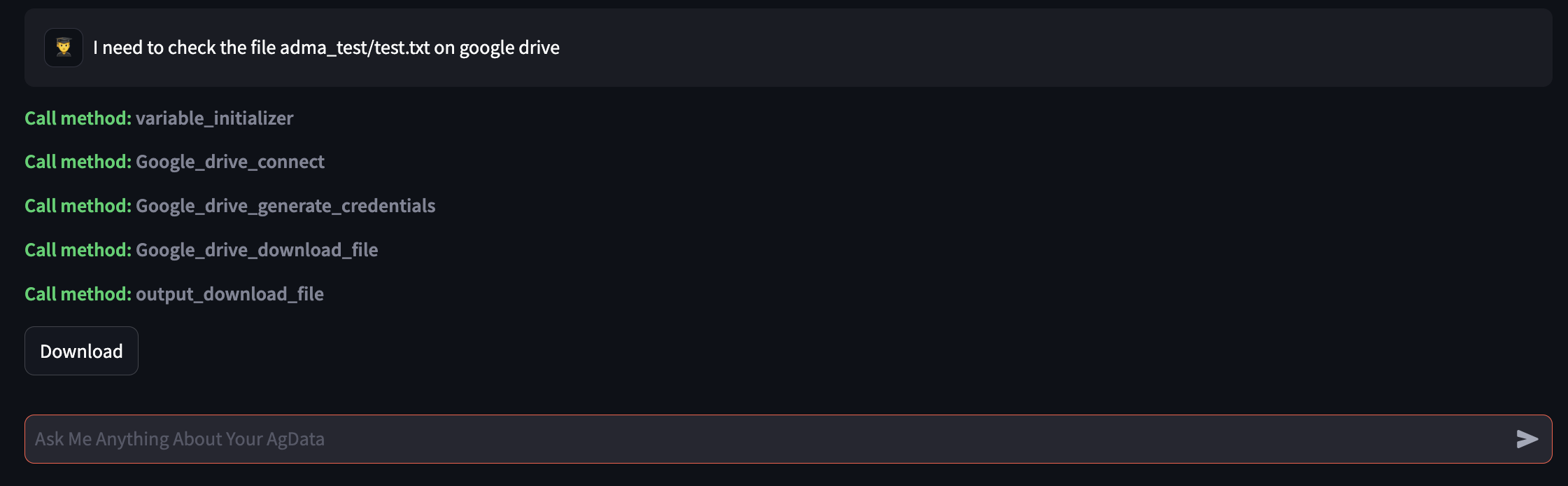}
        \caption{Check A File from Google Drive}
        \label{fig:demo_flexibility_2}
    \end{subfigure}%
    \hfill

    \caption{Demo of Flexible User Instruction}
    \label{fig:demo_flexibility}
\end{figure}

In Figure \ref{fig:demo_flexibility_1}, the user inputs "download the file adma\_test/test.txt on Google drive", then the copilot will download the specified file and display a download button on the interface. In Figure \ref{fig:demo_flexibility_2}, the user typs in  "I need to check the file adma\_test/test.txt on Google drive", the copilot can get the same intent as previous input and still display a download button on the interface.

\subsection{Privacy and Authentication}
\begin{figure}[t]
    \centering
    \begin{subfigure}{1\linewidth}
        \centering
        \includegraphics[width=\linewidth]{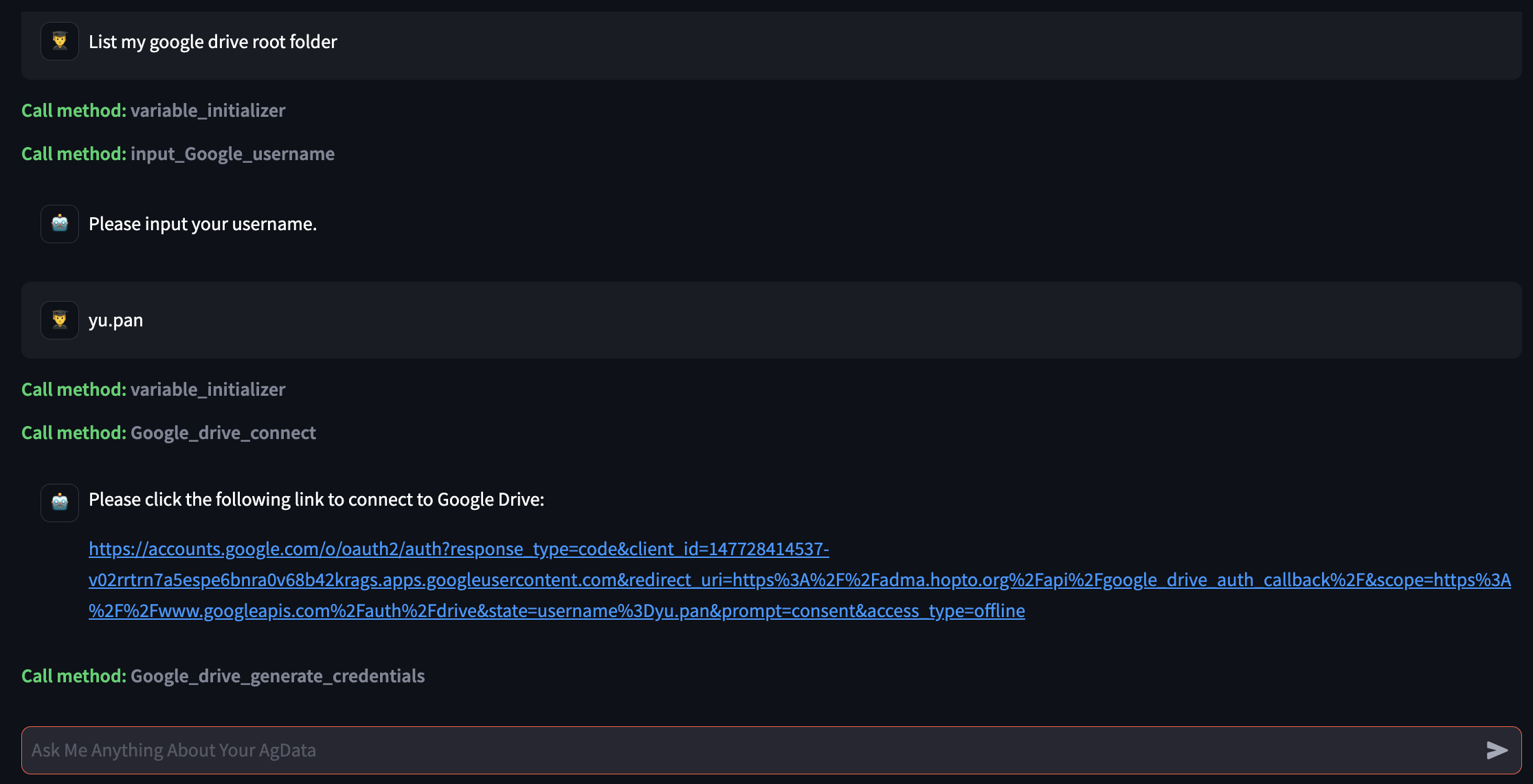}
        \caption{Ask for Google Credentials}
        \label{fig:demo_privacy_1}
    \end{subfigure}%
    \hfill
    \begin{subfigure}{1\linewidth}
        \centering
        \includegraphics[width=\linewidth]{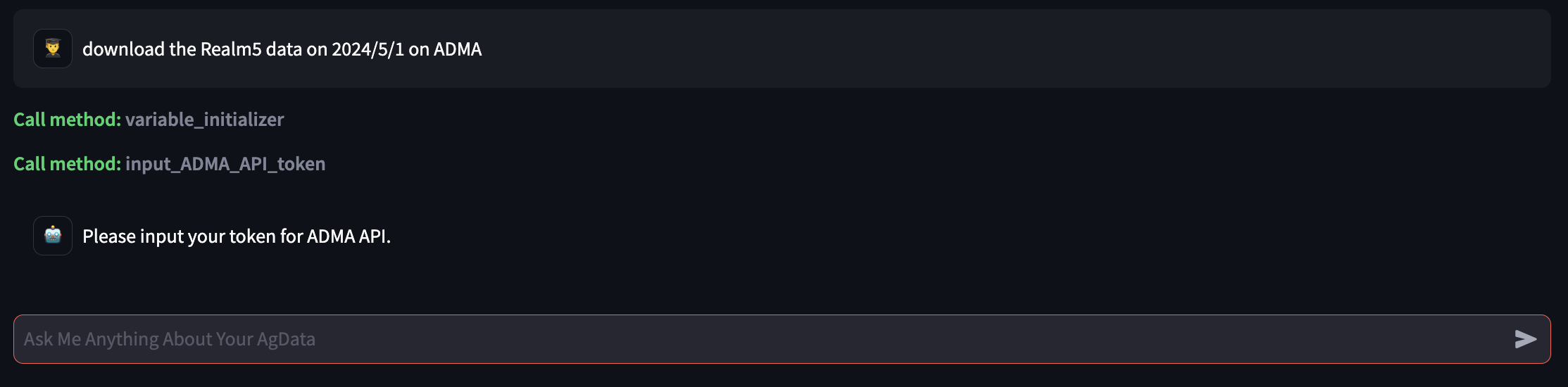}
        \caption{Ask for ADMA Token}
        \label{fig:demo_privacy_2}
    \end{subfigure}%
    \hfill

    \caption{Demo of Privacy Protection}
    \label{fig:demo_privacy}
\end{figure}

ADMA Copilot will protect the user's data and privacy, by ask for user's credentials to authenticate when the user need to get access to protected services or data. In Figure \ref{fig:demo_privacy_1}, the user want to "list my Google drive root folder", then the copilot decides it is a protected resource and ask for the user to go through the authentication process for Google drive. In Figure \ref{fig:demo_privacy_2}, the user types in "download the Realm5 data on 2024/5/1 on ADMA", then the copilot decides this is also a protected data and asks the user to input ADMA token.

\subsection{Comparison}

By adopting similar evaluation frameworks as in \cite{pan2023transforming}, we compare ADMA Copilot with the existing agriculture platforms. The evaluation criteria and corresponding explanation are listed in Table \ref{tab:criteria}. These criteria cover a broad range of dimensions, from the basic requirements for a typical data management platform to more innovative features, which also take into consideration the principles of FAIR. The eleven dimensions of criteria are, namely, privacy and security, CRUD repository, interoperability, extensibility, data model, data granularity, analytical tools, tractability, ML Models, and Intelligence and Autonomy. 

We make qualitative comparison between ADMA Copilot with existing agricultural data management platforms including CyVerse, GARDIAN, GEMS, TERRAREF and ADMA. Table \ref{tab:comparison} shows the comparison result. We mark $\times$ when a platform satisfies certain criteria and leave the cell empty otherwise.

ADMA Copilot meets all the criteria under consideration. While this comparison is qualitative and may introduce bias due to the selection and specific definition of evaluation criteria, the results suggest that our platform excels in the chosen dimensions. This indicates that the design principles of our system address a broader range of potential user needs and interests, positioning it well for future requirements.

\begin{table*}[htbp]
\caption{Dimensions of Evaluation Criteria}
\begin{center}
\begin{tabularx}{\linewidth}{|>{\hsize=.3\hsize}X|>{\hsize=1.7\hsize}X|}
\hline
\textbf{Dimension}&{\textbf{Explanation}} \\
\hline
Privacy\&Security  & The system protects data privacy and security, including but not limited to: 1. the ability to separate data and metadata; 2. the ability to share data or tools with other users safely; 3. the ability to differentiate public and private data; 4. the ability to run user-defined programs safely; 5. unified authentication module; 6. data encryption.\\
\hline
CRUD Repository & Data can be created (uploaded), read (rendered), updated, and deleted from/to/on the repository.  \\
\hline
Interoperability&  Data from external sources of different formats or various modalities can be added or incorporated into the system and managed by the system without difficulties.  \\
\hline
Extensibility &  New capabilities and functionalities can be added to the system. \\
\hline
Data Model &  Extract, transform and load procedure employed to populate a relational, graph, object, or vector database governed by an explicit data model. \\
\hline
Data Granularity & Ability to index and query for sub-file level entities such as individual records within datasets.\\
\hline
Analytical Tools &  Provides tools for users to clean text, numeric, and geospatial outliers, as well as conduct unique analysis.\\
\hline
Trackability & Ability to keep track of different operations on each file, maintain and visualize the pipeline record accordingly.\\
\hline
ML Models & Provide ready-to-use or support training and hosting of agricultural models.\\
\hline
Intelligence & The system utilizes Artificial Intelligence to facilitate data management, analysis, and interaction.\\
\hline
Autonomy & The system demonstrate certain degree of autonomy, by accomplishing data management pipeline with minimal interference or guiding from human users.\\
\hline
\end{tabularx}
\label{tab:criteria}
\end{center}
\end{table*}

\begin{table*}[t]
\caption{Comparison of Agriculture Data Management Platforms across a Variety of Criteria}
\begin{center}
\scalebox{0.72}{
\begin{tabular}{|c|c|c|c|c|c|c|c|c|c|c|c|}
\hline
\textbf{Platform} & \textbf{Privacy\& Security}& \textbf{CRUD Repository}& \textbf{Interoperability} & \textbf{Extensibility} & \textbf{Data Model} &\textbf{Data Granularity} & \textbf{Analytical Tools}& \textbf{Tractability} &\textbf{ML Models}& \textbf{Intelligence} & \textbf{Autonomy} \\
\hline
Cyverse & $\times$ & $\times$ & $\times$ & $\times$ &  & $\times$ & $\times$ &  & $\times$ & &\\
\hline
GARDIAN & $\times$ & $\times$ & $\times$ & $\times$ &  &  & $\times$ &  &  & &\\
\hline
GEMS & $\times$ & $\times$ & $\times$ & $\times$ &  &  & $\times$ &  & & &\\
\hline
TERRAREF & $\times$ & $\times$ & $\times$ & $\times$ & $\times$ & $\times$ & $\times$ &  & & &\\
\hline
ADMA & $\times$ & $\times$ & $\times$ & $\times$ & $\times$ & $\times$ & $\times$ & $\times$ & $\times$ & $\times$ &\\
\hline
ADMA Copilot& $\times$ & $\times$ & $\times$ & $\times$ & $\times$ & $\times$ & $\times$ & $\times$ & $\times$ & $\times$ &$\times$\\
\hline

\end{tabular}
}
\label{tab:comparison}
\end{center}
\end{table*}
\section{Conclusion}
\label{sec:conclusion}

In this paper, we propose and explore the idea of a LLM based multi-agent copilot for autonomous agricultural data management. Based on our previous developed platform: Agricultural Data Management and Analytics, we build a system called ADMA Copilot, which can understand user's intent, makes plan for data processing pipeline and accomplishes the task automatically, in which three agents: a LLM based controller, an input formatter and an output formatter collaborate together. Different from existing LLM based solutions, by defining a meta-program graph, our work decouples control flow and data flow to enhance the predictability of the behaviour of the agents. As a proof-of-concept product, ADMA Copilot can help to facilitate the shift of agricultural data management paradigm from traditionally human-driven paradigm to AI-driven paradigm. Experiments demonstrates the intelligence, autonomy, efficacy, efficiency, extensibility, flexibility and privacy of our system. Comparison is also made between ours and existing systems to show the superiority and potential of our system.

\section*{Acknowledgment}
This research is funded by the Nebraska Research Initiative. This work was completed utilizing the Holland Computing Center of the University of Nebraska, which receives support from the UNL Office of Research and Economic Development and the Nebraska Research Initiative.

\bibliographystyle{IEEEtran.bst}
\bibliography{IEEEabrv}

\end{document}